# Weakly Supervised Text Classification on Free Text Comments in Patient-Reported Outcome Measures

Anna-Grace Linton[1], Vania Dimitrova[2], Amy Downing[3], Richard Wagland[4], Adam Glaser[3]

[1]UKRI CDT in AI for Medical Diagnosis and Care, University of Leeds, UK
[2]School of Computing, University of Leeds, UK
[3]School of Medicine, University of Leeds, UK
[4]School of Health Sciences, University of Southampton, UK

**Abstract**

*Free text comments (FTC) in patient-reported outcome measures (PROMs) data are typically analysed using manual methods, such as content analysis, which is labour-intensive and time-consuming. Machine learning analysis methods are largely unsupervised, necessitating post-analysis interpretation. Weakly supervised text classification (WSTC) can be a valuable method of analysis to classify domain-specific text data in which there is limited labelled data. In this paper, we apply five WSTC techniques to FTC in PROMs data to identify health-related quality of life (HRQoL) themes reported by colorectal cancer patients. The WSTC methods label all the themes mentioned in the FTC. The results showed moderate performance on the PROMs data, mainly due to the precision of the models, and variation between themes. Evaluation of the classification performance illustrated the potential and limitations of keyword based WSTC to label PROMs FTC when labelled data is limited.*

## 1 Introduction

Patients' perspective of their health has become increasingly crucial when looking at the quality of survival of people who were diagnosed with cancer. It is considered key for a more holistic interpretation and understanding of their health conditions and their health-related quality of life (HRQoL) (1). Patient-reported outcome measures (PROMs) provide an assessment of a patient's HRQoL through a combination of closed questions, such as Likert scales, and open-ended questions (2). The responses to the open-ended questions in PROMs are presented usually as short free text comments (FTC). They can provide additional detail that complement the closed questions, allowing a more comprehensive understanding of the nuances of a patient's health and its influences (3,4).

While responses to closed questions in PROMs can be analysed efficiently using statistical methods, the analysis of free text responses is hard and such data is often left unexplored by clinical research (5–8). PROMs FTC analysis is significantly more time- and resource-demanding than closed question responses. This task is typically conducted manually by using qualitative analysis methods, which is prone to subjectivity, data-dependence, and does not scale. Often, FTC analysis is omitted in the analysis of PROMs, which can result in loss of information in reporting of the findings and potential bias (9,10). The demands of analysis are intensified by the increased use of patient reported data such as PROMIS and other PRO initiatives (11), and patient experience surveys collecting thousands free text responses each year which would take months to go through manually (12). The time it takes to analyse these comments can exceed the usefulness of the insights in the data.

Automated analysis of free text responses in PROMs can be enabled with the adoption of text analytics methods but brings key challenges. PROMs free text data is usually unlabelled. This data comes from patients in a free format and can relate to anything that patients want to raise about their quality of life. The unsupervised classification methods, which tend to be adopted in practical applications, offer solutions that are data-dependent and hard to generalise (13). A further challenge for finding appropriate text analytics methods to analyse PROMs FTC is the length of the data - both the individual contributions are short (around a paragraph), and the data sets are usually not very big (free text responses are optional in PROMs surveys and not all patients provide such responses).

A way to address these challenges is to adopt weakly supervised text classification (WSTC). Increasingly, WSTC is used when there is insufficient labelled data or it is costly to obtain expert annotations (14,15). WSTC uses weak supervision signals during training, such as keywords or heuristics, for text classification instead of labelled data (16). Consequently, the need for a large, annotated corpus can be avoided, which makes the approach quite appealing for

PROMs FTC data. Furthermore, keyword based WSTC can allow guidance from domain experts and thus can build on healthcare research on patients' quality of life. While WSTC is a promising approach, it has not been applied to PROMs FTC data, it is unclear how WSTC methods will perform on patient free text responses, and there is insufficient analysis as to whether WSTC is suitable for adoption in this specific text analytics context and in the broader context of healthcare.

In this paper, we investigate to what extent WSTC be adopted to enable automatic classification of patients' free text responses in PROMs data. We explore this in the context of free text data collected in an NHS colorectal cancer (CRC) PROMs survey – while the patients' responses to closed questions in that survey have been extensively analysed, the free text comments have not been as explored (17). This work is part of a PhD project, aimed at examining the value of free text comments in effectively and efficiently with minimal manual effort.

The paper presents a framework for utilising WSTC for classifying free text responses in a PROMs data set. Based on a scoping review (18), we have identified key themes related to health quality of life that could be presented in free text responses in PROMs together with the corresponding key words. Five keyword-based WSTC methods: BERTopic (19), CorEx Algorithm (20), Guided LDA (GLDA) (21), WeSTClass (22) and X-Class (23) have been applied to label free text data from the CRC PROMs survey with the predefined themes using seed terms representative of the themes. The performance of the algorithms is analysed. The insights were presented to the clinical research team discussing the feasibility of using the WSTC approach for PROMs FTC classification.

## 2 Relevant Work

**Analysing free text responses in PROMs.** Studies that analyse free text in patient-reported text data (including PROMs and patient-reported experience data) have been carried out using supervised and unsupervised approaches (13). Most automated approaches to analyse patients' free text responses are unsupervised, using information extraction (24–26); and classification (8,27–29). Maramba et al. (24) utilise several web-based text processing tools to extract useful information from free-text commentary Spasić et al., (25) mapped FTCs of knee osteoarthritis patients to the Likert scales of a PROMs dataset performing sentiment analysis and using MetaMap (30) to look up lexicon for named entity recognition. To analyse free text comments from an Irish in-patient survey, Robin et al. (26), used Saffron Software[1] to extract key terms in the medical domain and automatically mapped them to predefined categories. In both studies, the authors standardised and grouped responses using information extraction approaches to reduce human effort in analysis. Significant manual effort was required in providing annotated data to validate key word extraction methods. While these methods allow a keyword level analysis, thematic grouping is not conducted.

Several approaches have used unsupervised classification methods to derive the main themes in a corpus of free text responses. Wagland et al. (6) utilised unsupervised machine learning algorithms to identify the main themes of patient experiences, which allowed them to see the impact of care on health related quality of life, which was verified using qualitative analysis. Similarly, Arditi et al. (8) used text classification to derive main themes in the free text responses from the Swiss cancer patient experience survey. The derived themes related to personal and emotional experiences and consequences of having cancer and receiving care. Along the same line of research, Pateman et al. (28) utilised a text analytics tool to identify the main themes in free text patient responses about their experiences and quality of life outcomes in head and neck cancers. They extracted a concept map that identified main keyword clusters and linked them based on common terms. Sanders et al. combined text analytics and manual qualitative analysis to explore the usefulness of patient experience data in services for long-term conditions (29). They found that the comments gave meaning to otherwise meaningless quantitative scores, such as "neither likely or unlikely" and polarised experience. The authors argue that digital collection and automated analysis produced broad topics, but managed time and resource restrictions favourably compared to qualitative analysis. These methods show that grouping comments in themes is helpful for healthcare research. However, the findings from these methods are data dependent. Crucially, there is a need for additional human effort to analyse the themes and put meaningful labels, which can be prone to subjectivity.

In this paper we propose a supervised approach to identify the main themes in a corpus of patient FTC. A supervised approach for grouping comments was utilised by Rivas et al., (12) who developed a tool to automatically conduct thematic analysis to identify themes in a Welsh cancer patient experience survey. A rule-based information extraction was used and developed through co-design with healthcare researchers. The approach has the benefit of being able to be systematically applied to patient experience data to summarise the data. However, the extraction of rules required significant work and the themes were defined based on what was in the dataset, making it hard to transfer to another dataset without significant effort. Similarly to Rivas et al., our proposed framework aims to classify free text comments in pre-defined themes. However, we have derived the themes based on a scoping review of qualitative research that has analysed free text comments. Starting with the most common themes in PROMs, we have utilised weakly supervised short text classification methods to classify patients' free text responses in these pre-defined themes.

---

[1] https://github.com/insight-centre/saffron

**Weakly supervised short text classification.** Short text classification has gained interest due to the increase in generated short text such as social media, dealing with challenges such as ambiguity and data sparsity which makes information extraction difficult (31,32). Short text classification focusses on overcoming the challenges of classifying short text such as inadequate length and a shortage of word-occurrences resulting in ambiguity of the text due to lack of contextual information (31,33,34). Some methods look to enrich the contextual information for short text using external information such as knowledge bases (35,36). However, this requires the existence or creation of knowledge bases for that domain, which can require expertise and time to develop. Many short text classification methods require a large amount of annotated data, which is often not possible when dealing with free patient responses.

WSTC uses weakly supervised signals for text classification and overcomes the problem of small amounts of data (16). For classification it employs signals such as labelled documents (37,38), keywords representative of the class (39–43) or heuristics (14,44,45). These methods make it possible to automatically create training data rather than labelling data by hand. Keyword- or seed term-based WSTC has proven to be a popular approach to WSTC as users provide a set of keywords for each class, providing pseudo-labels or weak signals of the class. The set of keywords can be extensive or very short (42,46). Meng et al. (22), uses seed terms or class labels provided by the user as weak supervision. Their method generates pseudo documents to pretrain a neural classifier which is then refined through a self-training module with bootstrapping. Mekala and Shang, (41) use contextualised representations of a few human-provided seed words for pseudo-labelling of a contextualised corpus on two real world long text datasets. Gallagher et al., (20), used user provided seed terms to incorporate domain knowledge into CorEx to enable the guiding and interpretation of topics with "minimal human intervention". While WSTC shows promising results on various user-generated corpora, to the best of our knowledge. WSTC has not been applied to patient-reported text data. The aim of the work presented here is to investigate to what extent WSTC can be adopted for FTC in PROMs. We focus on keyword-based classification for short text as it allows domain experts to directly inject their domain knowledge in a low-cost manner, which is advantageous for the adoption of a method of analysis for PROMs and other healthcare text. We will provide a critical analysis of the feasibility of using WSTC for classifying FTC in PROMs.

## 3  Framework for free text PROMs classification

The main aim of our work is to develop a generic approach for analysing free text comments that can be adopted by health researchers to get deeper insights into PROMs. We propose a generic framework that can be applied to the patient reported comments in any PROMs questionnaire. The framework is presented in Figure 1.

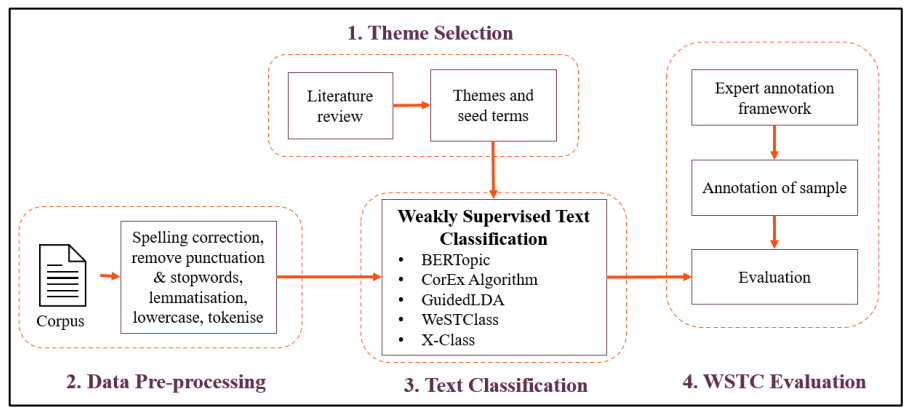

**Figure 1.** Framework for automated analysis of FTC in PROMs. The themes to identify in the free text comments are selected using themes found in a scoping review and refined by domain experts. The performance of five keyword based WSTC methods is evaluated on a CRC PROMs free text comments dataset.

**Theme selection.** We identified a set of reliable HRQoL themes reported by patients in FTC to label the comments in the PROMs data in a previously conducted scoping review (18). The themes were reported by patients with chronic conditions and grouped based on semantic similarity. For example, Study A called a theme "Fatigue" and Study B called a theme "Fatigue and Weakness", so both were recorded as two separate subthemes and grouped together as one theme. The themes were then categorised, aided by the WHO framework for HRQoL (47) - physical, psychological, independence, social and spiritual - for better understanding.

The grouping of these themes was reviewed, and grouping was agreed upon in concert with three domain experts. The subthemes and their prevalence were presented for explainability. The themes were again refined during the annotation of the data. A final list of themes was identified, aggregating similar themes, removing overlapping themes and adding missing themes. The final themes were: *Cancer Services & Pathways, Comorbidities, Physical Function, Psychological & Emotional Function, Social Function, and Daily Life*. Words and phrases associated with each theme were derived from the studies in the literature. They were used as the seed terms for this study. The seed terms acquired from the scoping review are used in this paper to guide the discovery of the themes within the PROMs FTC (see Table 1).

**Table 1.** The main themes identified from the review and the corresponding seed terms used for WSTC methods.

| Main Themes | Seed terms |
|---|---|
| Cancer Pathways & Services | ['radiotherapy', 'chemotherapy', 'surgery', 'treatment', 'diagnosis', 'diagnose', 'aftercare', 'referral', 'screen', 'monitoring', 'operation', 'stoma', 'nurse', 'doctor', 'staff', 'hospital'] |
| Comorbidities | ['angina', 'heart failure', 'copd', 'asthma', 'ulcer', 'stroke', 'dementia', 'parkinson', 'depression', 'melanoma', 'lymphoma', 'arthritis', 'old age', 'anxiety'] |
| Physical Symptoms | ['nausea', 'peripheral neuropathy', 'bleeding', 'cough', 'cold', 'fracture', 'vomit', 'sex life', 'sleep', 'weight', 'appetite', 'pain', 'ache', 'nausea', 'constipation', 'concern', 'diarrhoea', 'wind', 'constipation', 'bowel movement', 'fatigue', 'weakness', 'tiredness', 'energy', 'strength', 'memory', 'concentration', 'balance', 'mobility problem', 'sex'] |
| Psychological & Emotional Symptoms | ['embarrassment', 'fear', 'afraid', 'loss', 'worry', 'emotional', 'gratitude', 'praise', 'relief', 'hope', 'peace', 'faith', 'cop', 'pray', 'embarrass', 'cope', 'worried', 'confidence', 'mood', 'positive attitude', 'depressed', 'anxious', 'optimistic', 'attitude'] |
| Social Function | ['job', 'employment', 'financial', 'insurance', 'money', 'husband', 'wife', 'spouse', 'partner', 'grandchild', 'family', 'child', 'child', 'support family', 'social life', 'friend', 'community', 'dependent', 'socialise', 'isolation', 'isolated'] |
| Daily Life | ['travel', 'difficulty walk', 'usual activity', 'activity', 'lift', 'drive', 'diet', 'lifestyle', 'housework', 'exercise', 'active', 'physical activity', 'dress', 'hobby', 'wash', 'stairs'] |

**Data Pre-processing.** We used the Living With and Beyond Bowel Cancer survey PROMs dataset, which consisted of 5,634 responses that contained a free text response to an open-ended question for patients who had been diagnosed with colorectal cancer in the preceding 12-18 months at the time the survey was collected (17). Only 25.8% of all responders provided FTC. The length of comments ranged from 1 word to 269 words, with a mean of 43 words. The length of sentences ranged from 1 to 97 words (mean = 15 words). For CorEx, GLDA and WeSTClass, the comments were pre-processed to clean them. The comments had contractions expanded, lowercased, stop words removed, spelling corrected, and text tokenised. Documents containing one or fewer words were removed. The seed terms were processed in the same way. Term frequency-inverse document frequency was used as the word embedding for CorEx algorithm and GLDA. The processed text was provided to WeSTClass which produces a word embedding as part of the model. For X-Class and BERTopic, the raw text was provided as the input.

**Text Classification.** We apply five WSTC methods which have been applied and evaluated on short text.
- BERTopic – an embedding based method, that clusters documents based on a BERT based embedding. BERTopic is shown as advantageous as it provides continuous topic modelling rather than discrete (48). We use the Guided BERTopic version to initiate the topics that are formed.
- CorEx – a semi-supervised classification of the comments. This method finds maximally informative topics and through document correlation (20). Topics are "anchored" through provided seed terms.
- GLDA – uses class-related keywords as word-topic priors for supervision on a topic modelling-based method. This method is often used as a reliable baseline (21).
- WeSTClass - a neural network-based method which used a list of seed words to generate pseudo documents for self-training.

- X-Class – was chosen as it was shown as state-of-the-art for WSTC. Expanding provided surface labels to a list of seed terms, this method uses BERT embeddings to create pseudo-documents representative of each class and document-class pairs. These are then used to train a supervised model.

The parameters used were decided based on the performance of the methods overall a range of values. The methods are trained on the entire dataset (n=5635) and evaluated on a small sample (n=500). For all methods except for WeSTClass unigrams and bigrams are used as seed terms. For WeSTClass only unigrams are used as seed terms.

**WSTC Evaluation.** We used the framework described in (49) to prepare an annotated dataset to validate the performance of the WSTC methods. This included expert annotations to provide guidance and further annotation by two annotators.

Expert annotation. We first sampled 100 comments which were annotated by three experts with the six themes identified from the scoping review. The three experts (the last three authors of the paper) work in using PROMs to improve patient outcomes and were part of the past research team who collected the CRC PROMs data used in this paper. AG is a clincal oncologist who also uses PROMs for understanding the needs of individuals living with and beyond cancer. AD is a cancer epidemiologist and also researches the use of PROMs data for improving health practice and patient outcomes. RW is a health scientist with research in patient reported outcomes including free text comments.

The experts independently labelled the themes that were present in each comment and provide notes where applicable, such as on missing themes. They were provided with the comments, themes and related seed terms. The inter-annotator agreement was estimated for each theme using Krippendorff's Alpha (α) (50). Patient comments that had very high or low agreement were used as discussion prompts as well as patient comments that contained issues such as missing themes. As a consequence of the discussions, the themes were aggregated into broader (more generalised) themes, and human-understandable definitions were provided to the annotators. The experts were asked to annotate another 200 comments independently following the discussions in initial reviews of the themes. The agreement was calculated (**Table 2**). Moderately low agreement between the annotators suggests the subjectivity and ambiguity of the themes. As the agreement was at least moderate for all themes, majority voting between the annotators was used to provide a gold standard for the 200 comments. These annotated comments and definition refined by the expert annotators were used a framework for the annotators to follow.

**Table 2.** Agreement between three expert annotators on 200 comments) and prevalence of themes in labelled sample. A moderate agreement is 0.41 to 0.60 and a substantial is 0.61 to 0.80, and 0.81 and above is 'almost perfect' (50).

| | Themes | | | | | | |
|---|---|---|---|---|---|---|---|
| | Cancer Pathway & Services | Comorbidities | Physical Function | Psychological & Emotional Function | Social Function | Daily Life | No themes present |
| Agreement (α) | 0.79 | 0.728 | 0.653 | 0.605 | 0.635 | 0.608 | |
| Prevalence in sample (n=500) | 43% | 27% | 23% | 16% | 11% | 9% | 14% |

Sample Annotation. Two annotators annotated a larger sample of the comment for validation. These annotators were computing PhD students working in AI and PROMs data analysis. The annotators annotated the same 200 comments, and once agreement based on Cohen Kappa (51) was moderate (0.41 to 0.60) or substantial (0.61 to 0.80), they independently annotated an additional 300 new comments. The comments where there was disagreement was discussed, and the final labels were agreed. The result of the annotations was a sample of 500 comments labelled with the themes of HRQoL from the scoping review. The labels are not used during the training of the data but are only used to assess the performance of the WSTC methods. The prevalence of each theme in this sample is shown in **Table 2**. Each comment could contain more than one theme or no themes.

Methods' evaluation. To evaluate the WSTC methods' performance, we considered F1-score, recall, precision, and accuracy. We also carried out qualitative analysis of keywords extracted from the methods and labelled comments.

## 4. Experimental Results and Discussion

All 500 comments which have been annotated were used to evaluate the methods' ability to label the comments with HRQoL themes. Table 3 shows the example of two comments labelled by the methods and by the experts.

**Table 3.** Example comments that have been annotated and labelled by the WSTC methods.

| Patient Comment | Expert Annotation | BERTOPIC | CorEx | Guided LDA | WeSTClass | X-Class |
|---|---|---|---|---|---|---|
| I feel very fortunate and am thankful to all the hospital staff who treated me. | Cancer pathways & services | None | Cancer pathways & services | Cancer pathways & services | Psychological & Emotional Function | Cancer pathways & services |
| I am 86 yrs old my only problems bad circulation in both legs. I have had 2 varicose operations. I am my wife's carer as she has Alzheimer's Angina & Arthritis. Basically I am very fit and well for my age. 43 44 45 46 could alter dependent on my wife's condition. | Physical Function, Daily Life and Social Function | Physical function | Social function | Daily life | None | None |

**Method Performance.** As shown in Table 4 and Table 5, there is substantial variation in the performance of the methods with CorEx having the greatest overall accuracy (81.3%) and BERTopic having the lowest accuracy (32.1%). CorEx was the most accurate method for three themes (*Physical Function, Social Function* and *Daily Life*) while X-Class was the most accurate method for *Comorbidities*. Notably, BERTopic underperformed in all themes. The accuracy on *Cancer Pathway & Services* varied the least between methods (64.6 ± 7.1 STD). In contrast, *Daily Life* and *Social Function* were labelled with the greatest variation between methods. This can be expected as *Daily Life* and *Social Function* are more contextual and subjective themes compared to *Cancer Pathways & Services* (Table 2) and maybe contain more ambiguous concepts that some methods may or may not interpret well. The accuracy of the methods indicates their suitability for this data and the characteristics of the data for each theme.

**Table 4.** Mean accuracy score (+ standard deviation) across all themes for each method.

|  | Method | | | | |
|---|---|---|---|---|---|
|  | BERTopic | CorEx | GLDA | WeSTClass | X-Class |
| Mean (STD) | 32.1 (12.1) | 81.3 (7.2) | 74.2 (3.8) | 73.7 (3.6) | 74.7 (9.0) |

**Table 5**. Mean accuracy score for across all methods for each theme.

|  | Theme | | | | | |
|---|---|---|---|---|---|---|
|  | Comorbidities | Physical Function | Psychological & Emotional Function | Daily Life | Social Function | Cancer Pathway & Services |
| Mean (STD) | 65.2 (19.5) | 66.7 (18.5) | 69.1 (23.9) | 69.2 (26.3) | 68.4 (26.2) | 64.6 (7.1) |

We further explored the performance of the methods by reporting the F1-score, recall and precision of each model for each theme and observed, in many cases, large variations were because of precision (Figure 2). All methods performed the best on *Cancer Pathways & Services* (F1-score across methods = 0.60 – 0.72, Precision 0.8 - 0.9, Recall = 0.4 – 0.6). Performance for all methods was lowest for *Daily Life* (F1-score = 0.1-0.5, Precision = 0.1 – 0.4, excluding BERTopic which was 0.9, Recall = 0.1-0.7). GLDA and WeSTClass were comparable for all themes except for *Comorbidities* demonstrating relatively low precision for all themes except for *Cancer Pathway & services*. In contrast to the other methods, the precision for BERTopic was generally high across themes (Precision per theme = 0.78 - 0.87) but its performance was limited by very low recall (Recall per theme = 0.10 – 0.56). More conservative approaches that favour precision are desirable to ensure only relevant comments are labelled but risk producing a very narrow representation of the themes, with a large number of examples missed.

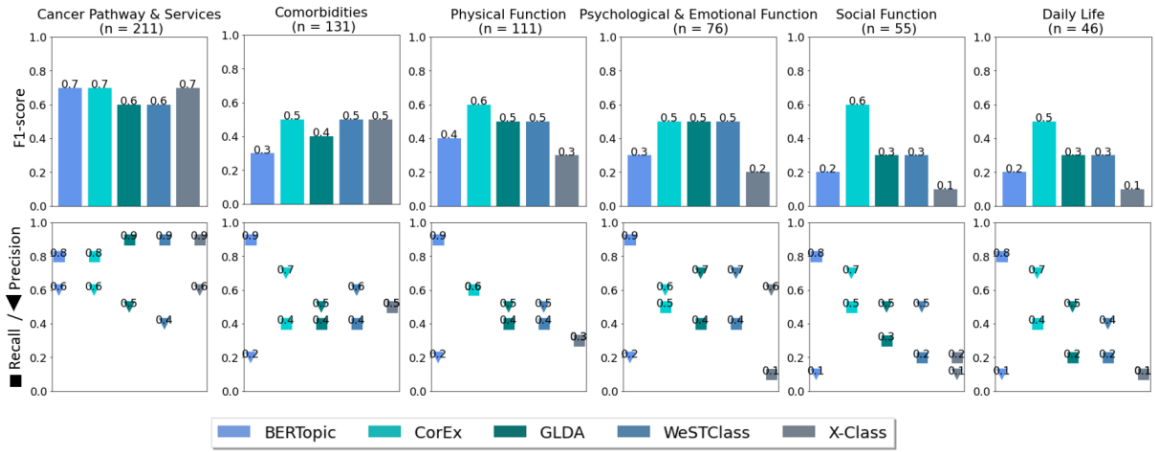

**Figure 2.** F1-score (top row), recall and precision (bottom row) of each method for each theme.

**Method Explainability.** Explaining how the methods work is crucial for the health research team to be able to assess the feasibility of the approach for FTC classification. To explain how the methods work, we extracted the keywords each method used for classification (Table 6). These were inspected by two of the domain experts, who identified for each method the words that could be relevant but were not used in the initial seed terms. The experts also commented on anything interesting they observed. Noticeably, some methods adhered strongly to the provided seed terms (WeSTClass) while others deviated greatly (BERTopic and X-Class). WeSTClass, unlike the other methods, found few other semantically relevant terms containing predominantly seed terms in the list of keywords, as shown by the words in bold. WeSTClass had a similar performance to GLDA, which captured relevant keywords and a lot of noise (irrelevant words shown in red). BERTopic differed from the seed terms the most and it failed to capture the seed terms in the top keywords for each theme. Other than some semantically relevant terms in *Comorbidities* such as "prostate cancer", the keywords mainly contained noise or terms relevant to other themes. For example, "chemotherapy" and "peripheral neuropathy" were captured in *Daily Life* instead of *Cancer Pathway & Services* and *Physical Function,* respectively. CorEx shows a mixture of the original seed terms and additional relevant keywords. *Comorbidities* was its noisiest theme with several words relating to CRC (the primary cancer), such as "bowel cancer" and "scan". A possible way to explain why CorEx performed reasonably well is that in addition to expert guidance (the initial seed terms), CorEx also looked at terms derived from the data and thus captured the patients' context.

The keywords were a good indication of what the methods learned to detect each theme and how well they did so. The seed terms provided are not extensive and intend to the range of concepts rather than to capture all the possible representations of a theme. Therefore, it was expected that methods that adhered too strongly to the seed terms would perform worse than the methods that were able to extrapolate and build upon the seed terms provided. Similarly, the methods that deviated excessively failed to generate themes of relevance.

Where the keywords are relevant, they demonstrated a disparity in experts' understanding and description of the themes and how patients discuss the theme in the free text comments, which impacted the performance of the methods. For example, "old (age)" was a seed term for *comorbidities* but was identified in GLDA in *Daily Life*. Whereas from clinicians' perspective, old age is considered a comorbidity, patients often describe aspects of their daily life that are affected because of old age. Another example is "thank", "grateful" and "excellent" which were captured as *Cancer Pathway & Services* by GLDA. Although these words are irrelevant to the theme, they indicate how the patients talk about *Cancer Pathway & Services* in the comments. The new keywords picked by the methods can give insight into how patients discuss and provide a clinically valuable context to the themes. As keyword-based WSTC rely heavily on the quality and relevance of seed terms to the task or dataset, a possible hybrid way to further refine and update the expert-driven seed terms with data-driven terms themes may improve precision scores for the methods (52).

**Table 6.** Keywords extracted from each method (n=15). The terms in bold are terms also in the seed terms provided. The terms in red were agreed by the domain experts as not relevant to the theme. The terms in blue are words that included in the themes by one expert with justifications.

| Theme | BERTopic | CorEx | GuidedLDA | WeSTClass | X-Class |
|---|---|---|---|---|---|
| Cancer Pathway & Services | questions, question, the questions, answers, page, answer, answered, some of, some, my answers, to answer, to, answers to, not | **hospital, nurse, staff, treatment, doctor, diagnosis**, care, excellent, screen, **aftercare, surgery**, receive, surgeon, district (nurse), monitoring | care, **treatment, hospital**, receive, excellent, **staff, nurse**, thank, surgeon, good, **doctor**, nhs, **diagnosis**, team, support, grateful | **nurse, doctor, hospital, radiotherapy chemotherapy, surgery, treatment, diagnosis**, diagnose, **aftercare**, referral, screen, monitoring, operation, stoma | **treatment**, treatments, therapy, treated, treating, treat, medication, intervention, care, **radiotherapy, chemotherapy**, medicine, **surgery**, clinical, **aftercare** |
| Comorbidities | bowel, cancer, bowel cancer, prostate, had, surgery, colon, was, prostate cancer, operation, in, diagnosed, cancer was, my, my bowel | cancer, bowel, bowel cancer, remove, liver, **arthritis**, lung, **stroke**, spread, scan, year, **depression**, ago, tumour, cancer spread | bowel, bowel cancer, scan, remove, liver, surgery, treatment, operation, diagnose, lung, month, year, test, check, chemotherapy, chemo | **angina, heart, diabetes, copd, asthma, ulcer stroke dementia parkinson depression**, melanoma, lymphoma, **arthritis**, old, **anxiety** | disease, illness, condition, disorder, syndrome, cancer, infection, failure, cancerous, problem, attack, dysfunction, malignant, symptom, tumour |
| Physical Function | thank, thank you, you, removed, staff, care, name, name removed, all, thanks, address, address removed, surgeon | **pain**, eat, **bowel movement, weight, diarrhoea, sleep**, foot, **peripheral neuropathy, wind, constipation**, tiredness, **energy, balance**, cold, **memory** | operation, bowel, problem, stoma, chemotherapy, day, hernia, time, surgery, month, reversal, cause, foot, leave, control | **nausea, neuropathy, bleeding, cough, cold, fracture, vomit, sleep, weight, appetite, pain**, ache, **nausea, constipation, diarrhoea** | symptoms, signs, complications, abnormalities, problems, conditions, issues, difficulties, spots, infections, attacks, effects, reactions, appeared, showing, |
| Psychological & Emotional Function | liver, lung, my liver, cancer, liver cancer, lungs, lung cancer, spread to, spread, secondary, 2012, the liver, bowel, had, liver and | **worry, hope, fear, loss, emotional**, feel, time, come, day, return, know, **faith**, think, thing, happen | support, nurse, information, need, help, patient, hospital, treatment, advice, helpful, surgery, follow, care, specialist, time | **Embarrassment, fear, afraid, loss, worry, emotional, gratitude, praise, relief, hope, peace, faith, cope, pray, embarrass** | psychological, emotional, mental, psychologically, emotionally, mentally, emotions, physically, physical, neurological, depression, feelings, **mood**, traumatic, memory, **anxiety**, |
| Social Function | removed name, name, name removed, removed, removed address, address removed, address, name and, my name, of name, number removed | **family, husband, wife, friend, insurance, job, support**, help, **financial, child**, life, **partner**, positive, make, die, **employment** | life, feel, live, help, positive, think, time, come, day, good, make, work, family, look, people | **Job, employment, family, community, insurance, money, husband, wife, spouse, partner, grandchild, child, social**, friend, dependent | social, **socialising**, public, **community**, personal, private, voluntary, group, society, practical, special, general, peoples, people, **friends** |
| Daily Life | chemotherapy, chemo, feet, hands, neuropathy, of chemotherapy, peripheral neuropathy, effects, and feet, of, of chemo, the chemotherapy, side | **diet, exercise, activity, travel, lifestyle, drive**, long, term, hernia, **walk**, long term, travel **insurance, housewook**, (colostomy) bag, difficulty walk | question, problem, answer, year, bowel, prostate, bowel cancer, age, ago, condition, mobility, prostate cancer, heart, relate, old | **Travel, walk, lift, drive, diet, lifestyle, housework, exercise**, active, **activity, dress, hobby, wash, stairs** | **lifestyle**, life, self, existence, living, live, everyday, healthy, normally, independent, normality, lead, **activity**, activities, **hobbies** |

## 5. Conclusions

The labelling of patient reported free text data is important to improve analysis and understanding of HRQoL and affecting factors from the perspective of patients. We looked at the extent to which five WSTC methods can be used to identify HRQoL in PROMs FTC, by applying the methods to CRC PROMs. Investigating the performance of the methods quantitatively using performance metrics and qualitatively using the keywords, allowed for comparison and interpretation, which is crucial for healthcare adoption. We found that the methods which could utilise expert-driven seed terms and learn contextually relevant words from the data performed the best overall. As consequence of the reduced effort for analysis, the adoption of free text PROMs can be encouraged. The next step for the work is to expand the number of comments used to evaluate the performance of the methods. To further explore the use of WSTC on patient reported data, future work will apply these methods to other patient-reported data, e.g. patient experiences.

**Acknowledgements.** The research reported here is part of a PhD project that has been funded by the UKRI Centre for Doctoral Training in Artificial Intelligence for Medical Diagnosis and Care (Project Reference: EP/S024336/1). This work uses data provided by patients and collected by a clinical researcher team as part of a research project on living with and beyond bowel cancer. We thank all the individuals who participated in this study and the National Cancer Registration Service.